\documentclass[letterpaper,times]{IONconf}

\usepackage{titling}
\usepackage{hyperref}
\usepackage{url}
\usepackage{bm}
\usepackage{fancyhdr}
\usepackage{courier}
\usepackage{amsmath}
\usepackage{amsfonts}
\usepackage[euler]{textgreek}
\usepackage{bbm}
\usepackage[]{graphicx}
\usepackage{float}
\usepackage{lipsum}
\usepackage{subcaption}
\usepackage{dsfont}
\usepackage{mathtools}
\usepackage{amssymb}
\usepackage{verbatim} 
\usepackage{multirow}
\usepackage{cite}
\usepackage[dvipsnames]{xcolor}

\usepackage[ruled,vlined,linesnumbered]{algorithm2e}
\IncMargin{0.7cm} 

\usepackage{siunitx}

\usepackage{outlines}
\usepackage{enumitem}
\setenumerate[2]{label=\alph*.}
\setenumerate[3]{label=\roman*.}

\newcommand{\state}{x}

\newcommand{\clkbias}{b}   
\newcommand{\clkbiasrate}{\dot{b}}   

\newcommand{\stdueremoon}{{\sigma_{\text{UERE, LNSS}}}}   
\newcommand{\stdlunareph}{\sigma_{\text{eph, LNSS}}}
\newcommand{\cno}{C/N_0}   
\newcommand{\speedoflight}{c}   

\newcommand{\stdlunarclk}{\sigma_{\text{clk, LNSS}}}   
\newcommand{\stdlunargd}{\sigma_{\text{gd, LNSS}}}   
\newcommand{\stdlunarrec}{\sigma_{\text{rec, LNSS}}}   

\newcommand{\crct}[1]{\hat{#1}}   
\newcommand{\predrate}{T_{\text{pred}}}   
\newcommand{\measupdaterate}{T_{\text{meas}}}   

\newcommand{\tidx}[1]{_{#1}}

\date{}
\title{\LARGE A Case Study Analysis for Designing a Lunar Navigation Satellite System with Time-Transfer from Earth-GPS}

\author{
    Sriramya~Bhamidipati, Tara~Mina and Grace~Gao, \textit{Stanford~University}
}

\begin{document}

\maketitle

\section*{biography}


\biography{Sriramya Bhamidipati}{is a postdoctoral scholar in the Aeronautics and Astronautics Department at Stanford University. She received her Ph.D. in Aerospace Engineering at the University of Illinois, Urbana-Champaign in 2021, where she also received her M.S in 2017. She obtained her B.Tech.~in Aerospace from the Indian Institute of Technology, Bombay in 2015. Her research interests include GPS, power and space systems, artificial intelligence, computer vision, and unmanned aerial vehicles.} 

\biography{Tara Mina}{is a Ph.D. candidate in the Electrical Engineering Department at Stanford University. 
She received her B.S. degree in Electrical Engineering from Iowa State University in 2017 and completed her M.S. degree in Electrical and Computer Engineering from the University of Illinois at Urbana-Champaign in 2019.}

\biography{Grace Gao}{is an assistant professor in the Department of Aeronautics and Astronautics at Stanford University.
Before joining Stanford University, she was an assistant professor at University of Illinois at Urbana-Champaign.
She obtained her Ph.D. degree at Stanford University.
Her research is on robust and secure positioning, navigation and timing with applications to manned and unmanned aerial vehicles, robotics, and power systems.}
\section*{Abstract}
Recently, there has been a growing interest in the use of a SmallSat platform for the future Lunar Navigation Satellite System~(LNSS) to allow for cost-effectiveness and rapid deployment. 
However, many design choices are yet to be finalized for the SmallSat-based LNSS, including the onboard clock and the orbit type. 
As compared to the legacy Earth-GPS, designing an LNSS poses unique challenges: (a) restricted Size, Weight, and Power~(SWaP) of the onboard clock, which limits the timing stability; (b) limited lunar ground monitoring stations, which engenders a greater preference toward stable LNSS satellite orbits.

In this current work, we analyze the trade-off between different design considerations related to the onboard clock and the lunar orbit type for designing an LNSS with time-transfer from Earth-GPS. 
Our proposed time-transfer architecture combines the intermittently available Earth-GPS signals in a timing filter to alleviate the cost and SWaP requirements of the onboard clocks. 
Specifically, we conduct multiple case studies with different grades of low-SWaP clocks and various previously studied lunar orbit types.
We estimate the lunar User Equivalent Range Error~(UERE) metric to characterize the ranging accuracy of signals transmitted from an LNSS satellite. 
Using the Systems Tool Kit (STK)-based simulation setup from Analytical Graphics, Inc. (AGI), we evaluate the lunar UERE across various case studies of the LNSS design to demonstrate comparable performance as that of the legacy Earth-GPS, even while using a low-SWaP onboard clock.
We further perform sensitivity analysis to investigate the variation in the lunar UERE metric across different case studies as the Earth-GPS measurement update rates are varied.


\section{Introduction}\label{sec:intro}

Over a half century since the end of the Apollo program, NASA will return humans to the Moon. 
In the coming decade, through the Artemis missions, National Aeronautics and Space Administration~(NASA) will land the first woman and person of color on the lunar south pole~\cite{smith2020artemis}. 
Additionally, many space agencies are participating in an international effort to establish a sustainable human presence on the Moon, which serves as a crucial platform to support future deep space exploration~\cite{laurini2014global}. 
Indeed, we are entering a second space race, with international space agencies planning over forty lunar missions in the next decade and with crucial involvement from the commercial space industry companies, including SpaceX and Blue Origin. 
To support the increasing plans for crewed and robotic activities, future lunar missions will require access to reliable and precise Position, Navigation, and Timing (PNT) services everywhere on the Moon. 


Recently, the NASA Goddard Space Flight Center (GSFC) and the European Space Agency~(ESA) have conceptualized GPS-like satellite constellations around the Moon, named LunaNet~\cite{israel2020lunanet} and Moonlight~\cite{cozzens2021moonlight}, respectively.
Furthermore, there has been an emerging interest in the use of a SmallSat platform for these PNT constellations to allow for cost-effectiveness and rapid deployment~\cite{israel2020lunanet}. 
These lunar PNT constellations by NASA and the ESA will assist in the overarching effort of establishing a sustainable human presence on the Moon, by providing global PNT and communication services to lunar users.
In particular, in the next decade, these initiatives seek to satisfy needs expressed by the Global Exploration community, with a targeted position accuracy of less than $50$~\si{m} for lunar users~\cite{cozzens2021moonlight}.

Given that the lunar PNT constellation initiatives are in the preliminary design phases, the design a Lunar Navigation Satellite System~(LNSS) involves finalizing many key design considerations, including:
\begin{itemize}
    \item Lunar satellite orbit. 
    Several types of lunar orbits that have previously been investigated include Low Lunar Orbit (LLO), Prograde Circular Orbit (PCO), Near-Rectilinear Halo Orbit (NRHO), and Elliptical Lunar Frozen Orbit (ELFO).
    In particular, ELFOs refer to a specific category of \textit{frozen orbits} providing a greater coverage of the lunar poles, wherein \textit{frozen} orbits represent the orbits that maintain nearly constant orbital parameters for extensive periods of time, without requiring station keeping~\cite{folta2006lunar, whitley2016options}. 
    Although PCOs are not frozen, they maintain multi-year stability with orbital parameters exhibiting predictable, repeatable behavior~\cite{whitley2016options}.
    NRHOs are highly elliptical orbits with nearly constant visibility of Earth and the lunar poles~\cite{schonfeldt2020system}.
    Due to the low orbiting altitude, LLOs have shorter orbital periods around the Moon, and there exist a few inclinations in which LLOs are also considered to be frozen or quasi-frozen~\cite{folta2006lunar}.
    \item Onboard clock. 
    The choice of onboard clock is critical for designing a navigation system as its grade (which depends on timing stability) directly affects the ranging precision offered to lunar users.
    Among various clock choices is the commercial Chip Scale Atomic Clock~(CSAC) with radiation-tolerance and low Size, Weight and Power~(SWaP), which has been specifically developed for space applications~\cite{schmittberger2020review}. Another potential clock choice is the Deep Space Atomic Clock~(DSAC), which has been recently designed by NASA to provide greater long-term timing stability and to assist in spacecraft radionavigation~\cite{ely2018using}.
\end{itemize}

Furthermore, designing a SmallSat-based LNSS involves unique challenges as compared to the legacy Earth-GPS, which lead to additional design limitations listed as follows: a)~Limited size of LNSS satellites. 
A SmallSat platform limits its payload capacity, including the SWaP of the onboard clock.
Given that lower SWaP clocks tend to have worse timing stabilities~\cite{schmittberger2020review}, the SWaP limitation on clock directly affects the timing stability;
b)~Limited ability to monitor LNSS satellites.
Given that a limited number of ground monitoring stations can be established on the Moon and that resources on Earth for monitoring the lunar constellation are limited, it is desirable for the LNSS satellites to require less maintenance, including fewer station-keeping maneuvers and clock correction maintenance; and
c)~Increased orbital perturbations in lunar environment. Because the Moon has a highly non-uniform distribution of mass~\cite{melosh2013origin}, its gravitation field is more anisotropic as compared to that of Earth's. 
In addition, Earth's gravity can significantly impact satellites in high-altitude orbits around the Moon, thereby limiting the set of feasible and stable lunar orbits. 


Given these challenges for designing a PNT constellation in the lunar environment, one may consider the potential in leveraging the existing Earth's legacy GPS, which is equipped with higher grade atomic clocks and an extensive ground monitoring network.
At lunar distances of about $385000$~\si{km}, the Earth-GPS signal is significantly attenuated, and the Earth-GPS satellites directed toward Earth are largely occluded by Earth and often the Moon. 
This limits the Earth-GPS signal availability at lunar distances to come from only the Earth-GPS transmit antenna's side lobes and the small, unoccluded parts of the main lobe. 
NASA’s magnetospheric multiscale mission, or MMS, has used these largely attenuated and intermittently available Earth-GPS signals to successfully compute position estimate in space~\cite{winternitz2017global}.
In fact, the MMS broke the Guinness World Record for the highest altitude of achieving an Earth-GPS fix in 2016 for traveling at distances of about one-fifth of the way to the Moon~\cite{johnson2016nasa,winternitz2017global} and then again in 2019 for distances of about halfway to the Moon~\cite{baird2019record}.  
Several simulated works also have demonstrated the feasibility of using Earth-GPS at lunar distances~\cite{schonfeldt2020system,winternitz2019gps,cheung2020feasibility}.
Through GPS Antenna Characterization Experiment~(ACE) study, NASA has characterized the GPS antenna gain patterns at high elevation angles from boresight for space users~\cite{donaldson2020characterization}. 
NASA has also developed a spaceborne Earth-GPS receiver~\cite{winternitz2004navigator}, which will be tested on the lunar surface for the first time in 2023~\cite{insideGNSS2021luGRE,kraft2020luGRE}.
Additionally, in 2023, the ESA will launch the Lunar Pathfinder communication satellite to the Moon, which will utilize a spaceborne, high-sensitivity Earth-GPS receiver to provide a position fix for the first time in lunar orbit~\cite{esa2021pathfinder}.

\begin{figure}[H]
\centering
\includegraphics[width=0.49\textwidth]{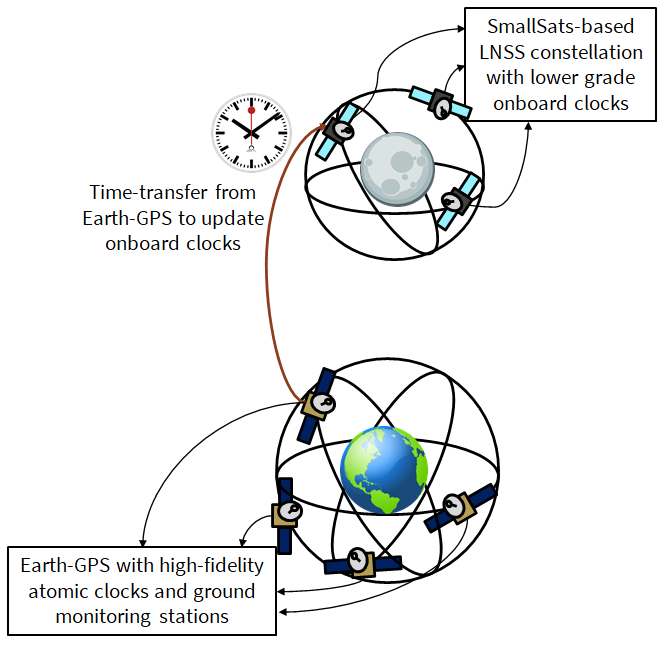}
\caption{Architecture of our proposed time-transfer from Earth-GPS~\cite{bhamidipati2021design}, which utilizes intermittently available Earth-GPS signals to correct the lower grade clocks onboard the LNSS satellite.}
\label{fig:proposedtimetransfer}
\vspace{-3mm}
\end{figure}

In our prior work~\cite{bhamidipati2021design}, we designed an LNSS architecture that harnessed the legacy Earth-GPS system to provide precise timing corrections to the onboard clock, as depicted in Fig.~\ref{fig:proposedtimetransfer}. 
Our proposed time-transfer technique leverages intermittently available Earth-GPS signals to alleviate the SWaP requirements of the onboard clocks and to mitigate the need for an extensive ground monitoring infrastructure on the Moon.
We also designed a lunar User Equivalent Range Error (UERE) metric, which is proportional to the Root-Mean-Square~(RMS) timing error, to analyze the ranging accuracy of an LNSS satellite.
We further demonstrated that our proposed method achieves a low UERE of less than $10$~\si{m} while using a low-SWaP Chip Scale Atomic Clock~(CSAC) for an LNSS satellite in an Elliptical Lunar Frozen Orbit~(ELFO).

Given that many design choices, including the grade of the onboard clock and the orbit type, still need to be finalized for the future SmallSat-based LNSS, in this work, we extend our prior work~\cite{bhamidipati2021design} to analyze the LNSS performance using our proposed time-transfer architecture~(in Fig.~\ref{fig:proposedtimetransfer}) from Earth-GPS under various case studies.
Specifically, we simulate an LNSS satellite in various lunar orbit types, including the ELFO, LLO, PCO, and NRHO, and equipped with different grades of onboard clocks. 
For a given LNSS satellite orbit, we examine an Earth-GPS Continual Outage Period~(ECOP) metric to analyze the visibility effects of Earth-GPS on the timing stability of the onboard clock. 
We further estimate the lunar UERE metric to perform comparison across different case studies. 
Through our analysis, we illustrate the trade-off between the different design considerations of the onboard clock and the orbit type for an LNSS design that leverages Earth-GPS time-transfer.
We additionally examine how the lunar UERE is affected by different rates of collecting available Earth-GPS measurements. 
This analysis provides insights into the extent of Earth-GPS signal tracking and processing required to provide sufficient ranging precision.
In particular, less frequent use of Earth-GPS measurements would allow the LNSS satellite to continually switch off the onboard Earth-GPS receiver for longer periods of time in order to save power.
Across the various case studies, we observe that with our time-transfer architecture, the LNSS can achieve comparable performance as that of the legacy Earth-GPS, even while using a low-SWaP onboard clock.

\subsection{Key Contributions} 
The key contributions of our paper are listed as follows:
\begin{enumerate}
    \item We design various case studies related to the grade of onboard clocks and the orbit types for analyzing the trade-offs in designing an LNSS with time-transfer from Earth-GPS.
    In particular, we investigate five clock types with diverse SWaP characteristics that range from a low-SWaP CSAC to a high-SWaP Deep Space Atomic Clock~(DSAC)~\cite{ely2018using} developed by NASA. 
    We also investigate four previously studied lunar orbit types, namely ELFO, NRHO, LLO and PCO. 
    \item We perform a comparison analysis across various case studies by investigating the associated RMS timing errors, which are in-turn governed by the duration for which no Earth-GPS satellites are visible~(ECOP metric) and the geometric configuration between the Earth-GPS constellation and the LNSS satellite~(occultations due to Earth and the Moon).  
    \item We evaluate the lunar UERE metric for each case study and demonstrate comparable measurement ranging accuracy as that of the legacy Earth-GPS, even while using a low-SWaP onboard clock. 
    \item We additionally examine how the lunar UERE is affected by different rates of collecting Earth-GPS measurements.
    This analysis provides insight into the extent of Earth-GPS signal processing required, and, correspondingly, the amount of power required to operate the onboard Earth-GPS receiver, in order to provide sufficient ranging precision
\end{enumerate}

The rest of the paper is organized as follows.
Section~\ref{sec:proposed_time_transfer_technique} summarizes our previously proposed time-transfer architecture from Earth-GPS to LNSS~\cite{bhamidipati2021design} and later describes the modifications incorporated in our time-transfer to conduct further analysis. 
Section~\ref{sec:simulation_modeling} provides a high-level overview of various case studies and also describes our high-fidelity lunar simulation setup that involves modeling the onboard clock and the orbit for each case study. 
Section~\ref{sec:experiment_results} discusses the implications of our case study analysis in designing an LNSS. 
Section~\ref{sec:conclusion} provides concluding remarks.
\section{Our Time-Transfer from Earth-GPS to LNSS} \label{sec:proposed_time_transfer_technique} 

In this section, we summarize our prior work~\cite{bhamidipati2021design} on time-transfer from Earth-GPS, wherein, we consider an LNSS satellite equipped with a Earth-GPS receiver and an onboard clock that can provide short-term timing stability.
We designed a timing Kalman filter~\cite{krawinkel2016benefits, martoccia1998gps} to update the LNSS satellite clock with the intermittently available Earth-GPS signals and formulated the lunar UERE metric to characterize the ranging accuracy of transmitted navigation signals.
We additionally provide an overview on the aspects of further analysis conducted in this work.
In particular, we analyze the sensitivity of the lunar UERE metric in different simulated case studies, including modifications of the measurement update rate for our proposed timing Kalman filter. 

At any LNSS satellite, our proposed filter maintains the LNSS clock estimate at each time epoch~$t$ by propagating the following timing state vector: $\state\tidx{t} \coloneqq \begin{bmatrix} \clkbias\tidx{t} & \clkbiasrate\tidx{t} \end{bmatrix}^{\top}$, where $\clkbias\tidx{t}$ is the clock bias state in \si{m} and $\clkbiasrate\tidx{t}$ is the clock drift in \si{ms^{-1}}, with units converted from the timing domain through multiplication by the speed of light $\speedoflight=299792458$~m/s.

To maintain the LNSS clock estimate, our timing Kalman filter performs a \textit{time update} every $\predrate$ seconds, based on the clock error propagation model.
For this, we define the associated process noise covariance $Q$ in terms of the power spectral density coefficients $h_0$, $h_{-1}$, $h_{-2}$ from the clock Allan deviation plot~\cite{krawinkel2016benefits}.
These Power Spectral Density~(PSD) coefficients reflect the short-term and long-term stability of the onboard clock~\cite{van1984relationship}.
To perform time-transfer from Earth-GPS, we first determine if any Earth-GPS signals are visible by examining the received carrier-to-noise-density ratio~$\cno$.
Then, our timing Kalman filter conducts a \textit{measurement update} for the available Earth-GPS measurements with sufficiently high $\cno$.

During the measurement update step, we determine the expected pseudorange and pseudorange rates from the visible Earth-GPS satellites to form a measurement vector of residuals.
In particular, we formulate the measurement residual vector by leveraging the position aiding from available ephemeris information for the LNSS satellite. 
We additionally model the measurement covariance matrix as a time-dependent diagonal matrix~\cite{bhamidipati2021design}, based on the tracking errors of the receiver Delay Lock Loop~(DLL)~\cite{kaplan2017understanding,capuano2015feasibility,capuano2016gnss} and Phase Lock Loop~(PLL)~\cite{borio2011doppler, capuano2016gnss} as well as the Earth-GPS UERE~\cite{kaplan2017understanding} and the expected error in available LNSS satellite ephemeris.
With our measurement vector and modeled measurement covariance, our filter applies corrections to the predicted timing state via the standard Kalman filter expressions to obtain the updated state $\crct{\state}\tidx{t}$ and covariance.

Based on the RMS error in our filter estimate, we formulate a lunar UERE metric that characterizes the accuracy of the LNSS ranging signals for lunar users.
On the Moon, any atmospheric delays are minimal. Moreover, we consider multipath effects experienced by users on the lunar surface to be negligible due to the lack of building infrastructure and foliage. 
As a result, the final lunar UERE can be computed in terms of four most significant error components as follows:
\begin{align}
    \stdueremoon = \sqrt{\stdlunarclk^2 +\stdlunargd^2+\stdlunareph^2+\stdlunarrec^2},
\end{align}
where the errors due to the differential group delay~$\stdlunargd$ and the receiver noise~$\stdlunarrec$ will depend on the final LNSS signal structure and lunar user receiver. Note that, because the timing filter uses position aiding from the LNSS satellite, the lunar ephemeris error component~$\stdlunareph$ directly impacts the pseudorange residual measurement received at the LNSS, which will thus also affect the LNSS clock error~$\stdlunarclk$.

In our current work, we investigate how the LNSS clock error component~$\stdlunarclk$ differs for various grades of onboard clocks and for various types of lunar orbits, while also analyzing the corresponding impact on the overall lunar UERE.
We additionally investigate the impact to the lunar UERE for a potentially reduced measurement update rate when Earth-GPS signals are available, with a sampling period of $\measupdaterate = m\predrate$ seconds, where $m$ is a positive integer.
Indeed, a larger choice of $m$ corresponds to less frequent GPS measurement updates, which allows the spaceborne Earth-GPS receiver onboard an LNSS satellite to be switched off for longer durations of time to save power. 
\section{Overview of Case Studies on Clocks and Orbits} \label{sec:simulation_modeling}

We perform an extensive case study analysis to examine the trade-off between different choices of the onboard clocks and the orbit types that can be considered for designing an LNSS with time-transfer from Earth-GPS. 
In particular, we design a high-fidelity simulation of an LNSS satellite for each orbit type using the Systems Tool Kit~(STK) software by the Analytical Graphics, Inc.~(AGI)~\cite{stk}. 
For each modeled orbit type, we design case studies in MATLAB by simulating various grades of onboard clocks. 
For each case study, we consider the start time epoch to be $9$ Nov $2025$ $00:00:00.000$ UTC and the experiment time duration to be $2$ months~(which equals $61$~days).

We first provide an overview of case studies related to the onboard clocks and the lunar orbit types investigated in this work. 
Thereafter, we summarize the simulation steps executed in the STK software and MATLAB for modeling the transmission and reception of Earth-GPS signals in each case study, which is based on the validation framework designed for our earlier work~\cite{bhamidipati2021design}. 

As mentioned in Section~\ref{sec:intro}, many prior studies~\cite{delepaut2020use,schonfeldt2020across} have investigated various types of lunar orbits, primarily based on their stability and the duration for which their stability can be ensured. 
Our choice of lunar orbit types for this work include ELFO, NRHO, LLO and PCO.
We create realistic simulations of an LNSS satellite in different lunar orbits by leveraging the High Precision Orbit Propagator~(HPOP) in the STK software~\cite{stk}. 
The HPOP generates and propagates accurate position and velocity solutions of the LNSS satellite by accounting for precise force models of Earth, the Sun and the Moon~\cite{chao2000independent}.

\begin{table}[H]
\centering
\caption{Keplarian parameters of the three~(ELFO, LLO and PCO) among the four lunar orbit types considered in our case study analysis.}
\begin{tabular}{|c|c|c|c|c|c|c|}
\hline
\textbf{\begin{tabular}[c]{@{}c@{}}Orbit Type \\ \end{tabular}} & \textbf{\begin{tabular}[c]{@{}c@{}}Altitude \\ (\si{km})\end{tabular}} & \textbf{Eccentricity} & \textbf{\begin{tabular}[c]{@{}c@{}}Inclination\\ ($^{\circ}$)\end{tabular}} & \textbf{\begin{tabular}[c]{@{}c@{}}Argument of\\ Perigee ($^{\circ}$)\end{tabular}} & \textbf{\begin{tabular}[c]{@{}c@{}}RAAN \\ ($^{\circ}$)\end{tabular}} & \textbf{\begin{tabular}[c]{@{}c@{}}Mean Anomaly\\ ($^{\circ}$)\end{tabular}} \\ \hline
\textbf{ELFO} & 9750.5 & 0.7 & 63.5 & 90 & 0 & 0 \\ \hline
\textbf{LLO} & 100 & 0 & 28.5 & 0 & 0 & 0 \\ \hline
\textbf{PCO} & 3000 & 0 & 75 & 0 & 90 & 0 \\ \hline
\end{tabular}
\label{table:LNSSsatorbits}
\end{table}

\begin{figure}[H]
\centering
\includegraphics[width=0.45\textwidth]{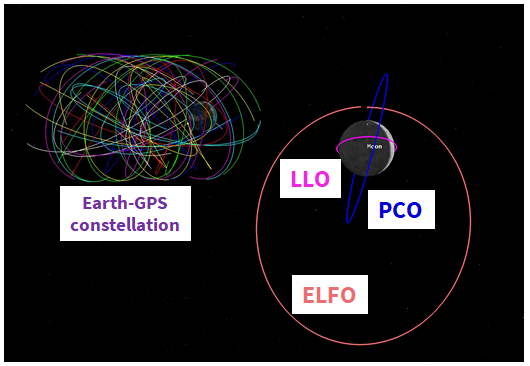}
\caption{Illustration of the three~(ELFO, LLO and PCO) among the four lunar orbit types considered in our case study analysis.
All these orbits are visualized in the Moon-inertial frame, where ELFO indicated in orange is at an altitude of $9750.5$~\si{km}, LLO represented in magenta is at $100$~\si{km}, and PCO represented in blue is at $3000$~\si{km}.}
\label{fig:LNSSsatelliteThreeOrbitTypes}
\end{figure}

We model the three orbit types, namely ELFO, LLO and PCO, in the STK software using classical orbit mechanics~\cite{montenbruck2012satellite}, according to which objects orbiting in space requires six elements (in our case six Keplarian parameters) to fully characterize their position and velocity at any point in time.
Specifically, we refer to prior literature~\cite{ely2006constellations,delepaut2019system,whitley2016options} on ELFO, LLO and PCO to define their corresponding six Keplarian parameters at the start time epoch, which includes semi-major axis, eccentricity, inclination, argument of perigee, Right Ascension of the Ascending Node~(RAAN), and mean anomaly. 
Table~\ref{table:LNSSsatorbits} lists the associated Keplarian parameters of the three lunar orbit types, while Fig.~\ref{fig:LNSSsatelliteThreeOrbitTypes} shows their associated illustration in the Moon-inertial frame. 

\begin{figure}[H]
	\centering
	\begin{subfigure}[b]{0.49\textwidth}
		\includegraphics[width=0.92\textwidth]{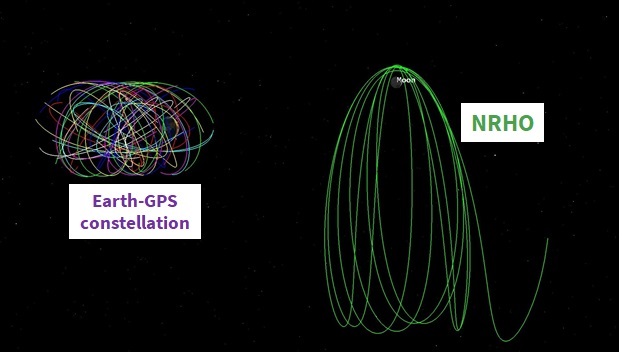}
		\caption{NRHO orbit type in Moon-inertial frame}
		\label{fig:LNSSsatelliteNRHOMoonIntertial}
	\end{subfigure}	
	\hspace{2mm}
	\begin{subfigure}[b]{0.43\textwidth}
		\includegraphics[width=0.98\textwidth]{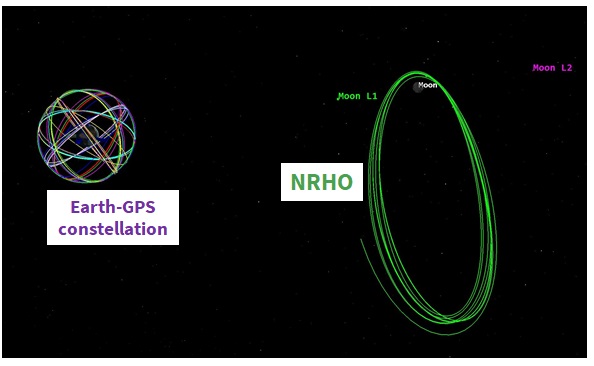}
		\caption{NRHO orbit type in Moon-centered Earth-Moon rotating frame}
		\label{fig:LNSSsatelliteNRHOEarthMoonRotating}
	\end{subfigure}
	\caption{The NRHO~(L2, radius of perigee=$4500$~\si{km}, South family) visualized in (a)~the Moon-Inertial frame; and (b)~the Moon-Centered Earth-Moon rotating frame.
	The x-axis of the Earth-Moon rotating frame is along the instantaneous Earth-Moon position vector, the z-axis is along the instantaneous angular momentum vector of the Moon’s orbit around the Earth, and the y-axis completes the orthogonal system.} 
	\label{fig:LNSSsatelliteNRHO}
\end{figure}

In contrast, for modeling the fourth lunar orbit type in our case study analysis, namely L2 South NRHO, we specify the initial conditions of position~$[r_x, r_y, r_z]$, and velocity~$[v_x, v_y, v_z]$ in the Moon-centered Earth-Moon rotating frame. 
Defining the initial conditions in an Earth-Moon rotating frame is particularly useful when discussing halo orbits in the Earth-Moon system~\cite{williams2017targeting}. 
In the Earth-Moon rotating frame, the x-axis is along the instantaneous Earth-Moon position vector, the z-axis is along the instantaneous angular momentum vector of the Moon’s orbit around the Earth, and the y-axis completes the orthogonal system.
Specifically, we refer to prior literature~\cite{williams2017targeting} that solves for an NRHO in the ephemeris model~(L2, radius of perigee to be $4500$~\si{km}, South family) using a forward/backward shooting process to provide the following initial state vector at an initial time epoch of $8$ Nov $2025$ $23:22:07.10353~$TDB:~$r_x=-125.952$~\si{km}, $r_y=120.961$~\si{km}, $r_z=4357.681~$km, $v_x=-0.042$~\si{km/s}, $v_y=1.468$~\si{km/s} and $v_z=-0.003$~\si{km/s}. 
The illustrations of the designed NRHO both in the Moon-inertial frame~(similar frame as that in Fig.~\ref{fig:LNSSsatelliteThreeOrbitTypes}) and in the Moon-Centered Earth-Moon rotating frame are shown in Figs.~\ref{fig:LNSSsatelliteNRHOMoonIntertial} and~\ref{fig:LNSSsatelliteNRHOEarthMoonRotating}, respectively.

Given the interest in SmallSat platform for the future LNSS, we choose various case studies of onboard clock types while keeping in mind that the limited payload capacity restricts the SWaP of the onboard clock.
Based on prior literature~\cite{schmittberger2020review}, our choice of clock types for this work include Microchip Chip Scale Atomic Clock~(CSAC), Microchip Micro Atomic Clock~(MAC), Stanford Research Systems~(SRS) PRS-10, Excelitas Rubidium Atomic Frequency Standard~(RAFS) and NASA's Deep Space Atomic Clock~(DSAC)~\cite{ely2018using}.
The specifications of these clock types are listed in Table~\ref{table:LNSSsatclocks}, which have been arranged in the increasing order of their SWaP for convenience.
For each clock type, we simulate the true clock error model in MATLAB to have a constant drift in the clock bias and thereafter, propagate the clock states forward in time using a first-order state transition matrix. 
We assign the true clock drift~(constant value) based on the known specifications of Time Deviation~(TDEV) observed at the end of a day, which are reported in Table~\ref{table:LNSSsatclocks}. 
Note that, for any clock type, TDEV refers to the expected error in reported time after a certain holdover time that essentially depends on their respective values of Allan deviation and frequency drift. 
For \textit{time update} in our timing filter described previously in Section~\ref{sec:proposed_time_transfer_technique}, we utilize the PSD coefficients listed in Table~\ref{table:LNSSsatclocks} to model the corresponding process noise covariance matrix~$Q$.

\begin{table}[H]
\centering
\caption{SWaP characteristics, time deviation and PSD coefficients of the five clock types considered in our case study analysis. 
SWaP and TDEV values are taken from~\cite{schmittberger2020review} while PSD coefficients were computed from their respective Allan deviation plots.}
\begin{tabular}{|c|c|c|c|c|ccc|}
\hline
\multirow{3}{*}{\textbf{Clock type}} & \multirow{3}{*}{\textbf{\begin{tabular}[c]{@{}c@{}}Size\\ (\si{cm^3})\end{tabular}}} & \multirow{3}{*}{\textbf{\begin{tabular}[c]{@{}c@{}}Weight\\ (\si{kg})\end{tabular}}} & \multirow{3}{*}{\textbf{\begin{tabular}[c]{@{}c@{}}Power\\ (\si{W})\end{tabular}}} & \multirow{3}{*}{\textbf{\begin{tabular}[c]{@{}c@{}}TDEV per day\\ (\si{\mu s})\end{tabular}}} & \multicolumn{3}{c|}{\textbf{PSD coefficients}} \\ \cline{6-8} 
 &  &  &  &  & \multicolumn{1}{c|}{\multirow{2}{*}{$h_{0}$}} & \multicolumn{1}{c|}{\multirow{2}{*}{$h_{-1}$}} & \multirow{2}{*}{$h_{-2}$} \\
 &  &  &  &  & \multicolumn{1}{c|}{} & \multicolumn{1}{c|}{} &  \\ \hline
\multirow{2}{*}{\textbf{Microchip CSAC}} & \multirow{2}{*}{17} & \multirow{2}{*}{0.035} & \multirow{2}{*}{0.1} & \multirow{2}{*}{1.5$\times10^{-6}$} & \multicolumn{1}{c|}{\multirow{2}{*}{1.3$\times 10^{-20}$}} & \multicolumn{1}{c|}{\multirow{2}{*}{1.0$\times10^{-24}$}} & \multirow{2}{*}{3.7$\times10^{-29}$} \\
 &  &  &  &  & \multicolumn{1}{c|}{} & \multicolumn{1}{c|}{} &  \\ \hline
\multirow{2}{*}{\textbf{Microchip MAC}} & \multirow{2}{*}{50} & \multirow{2}{*}{0.084} & \multirow{2}{*}{5} & \multirow{2}{*}{1.7$\times 10^{-7}$} & \multicolumn{1}{c|}{\multirow{2}{*}{4.7$\times 10^{-22}$}} & \multicolumn{1}{c|}{\multirow{2}{*}{1.2$\times 10^{-25}$}} & \multirow{2}{*}{1.7$\times10^{-30}$} \\
 &  &  &  &  & \multicolumn{1}{c|}{} & \multicolumn{1}{c|}{} &  \\ \hline
\multirow{2}{*}{\textbf{SRS PRS 10}} & \multirow{2}{*}{155} & \multirow{2}{*}{0.6} & \multirow{2}{*}{14.4} & \multirow{2}{*}{7.0$\times 10^{-8}$} & \multicolumn{1}{c|}{\multirow{2}{*}{1.3$\times10^{-22}$}} & \multicolumn{1}{c|}{\multirow{2}{*}{2.3$\times10^{-26}$}} & \multirow{2}{*}{3.3$\times10^{-31}$} \\
 &  &  &  &  & \multicolumn{1}{c|}{} & \multicolumn{1}{c|}{} &  \\ \hline
\multirow{2}{*}{\textbf{Excelitas RAFS}} & \multirow{2}{*}{1645} & \multirow{2}{*}{6.35} & \multirow{2}{*}{39} & \multirow{2}{*}{4.8$\times10^{-9}$} & \multicolumn{1}{c|}{\multirow{2}{*}{8.0$\times10^{-24}$}} & \multicolumn{1}{c|}{\multirow{2}{*}{0}} & \multirow{2}{*}{0} \\
 &  &  &  &  & \multicolumn{1}{c|}{} & \multicolumn{1}{c|}{} &  \\ \hline
\multirow{2}{*}{\textbf{NASA DSAC}} & \multirow{2}{*}{17000} & \multirow{2}{*}{16} & \multirow{2}{*}{47} & \multirow{2}{*}{4.0$\times10^{-11}$} & \multicolumn{1}{c|}{\multirow{2}{*}{1.8$\times10^{-27}$}} & \multicolumn{1}{c|}{\multirow{2}{*}{0}} & \multirow{2}{*}{0} \\
 &  &  &  &  & \multicolumn{1}{c|}{} & \multicolumn{1}{c|}{} &  \\ \hline
\end{tabular}
\label{table:LNSSsatclocks}
\end{table}

For a given LNSS satellite clock and orbit type in each case study, we model the simulation scenario in the STK software and MATLAB to compute the following statistics that are later given as input to our timing filter: the $\cno$ and measurement residual vector for visible Earth-GPS satellites.
We summarize the key modeling aspects below, while a more detailed explanation regarding the same has been discussed in our earlier work~\cite{bhamidipati2021design}. 

First, we design our simulated Earth-GPS constellation to be made up of $31$ satellites with $8$ satellites from Block IIR, $7$ from IIRM, $12$ from IIF and $4$ from Block III. 
We model the transmit antenna of Earth-GPS satellites by utilizing the transmit power and antenna gain patterns of the L1~C/A signals, which are made available from the NASA GPS ACE study~\cite{donaldson2020characterization}. 
Next, we simulate a spaceborne Earth-GPS receiver with a steering antenna pointed towards the Earth so as to maximize the visibility of Earth-GPS signals at the LNSS satellite. 
Based on prior literature~\cite{delepaut2019system,capuano2016gnss}, we consider a high-gain antenna with $14$~\si{dBi} at $0^{\circ}$ off-boresight angle, and a $3~$dB beamwidth of $12.2^{\circ}$. 
We consider an Earth-GPS satellite to be visible when the received $\cno$ value is greater than $15$~\si{dB-Hz} for a continuous time duration of at least $40$~\si{s}.

Finally, we simulate received measurements at the LNSS satellite by incorporating the true clock bias and drift in the true range and range rate between the visible Earth-GPS and LNSS satellites, respectively. 
Note that the $\cno$, true range and range rate values are extracted from the STK simulation while the true clock bias and drift are obtained from the simulated clock error model in MATLAB.
To formulate the residual vector given to \textit{measurement update} explained previously in Section~\ref{sec:proposed_time_transfer_technique}, we induce stochastic errors based on the simulated uncertainties from the receiver tracking loops. 
\section{Our Case Study Analysis: Results and Discussion} \label{sec:experiment_results}
For designing an LNSS with time-transfer from Earth-GPS, we analyze the trade-off across different case studies related to the onboard clock and the orbit type, which were listed earlier in Section~\ref{sec:simulation_modeling}.

To characterize the lunar UERE discussed in Section~\ref{sec:proposed_time_transfer_technique}, we consider the same group delay and receiver noise error magnitudes as with the Earth-GPS system, i.e. $\stdlunargd = 0.15$~\si{m} and $\stdlunarrec=0.1$~\si{m}.
Given that the future LNSS will have greater limitations on ground monitoring infrastructure as compared to Earth-GPS, we perform case study analysis by scaling the error component due to the broadcast ephemeris in the lunar UERE as $\stdlunareph=3$~\si{m}. 
This value is essentially one order of magnitude higher in comparison to that of Earth-GPS.  
We design our timing filter with the \textit{time update} executed every $\predrate=60$~\si{s}. 
For the \textit{measurement update}, details regarding the reduced update rate with a sampling period of $\measupdaterate=m\predrate$ for \textit{measurement update} will be discussed below.

\subsection{Validation Metrics} 
To perform comparison analysis across different case studies, we define the following four validation metrics: 
\begin{enumerate}
    \item Satellite visibility, which indicates the percentage time in the entire experiment duration for which the number of visible Earth-GPS satellites are greater than a pre-specified threshold. 
    We examine two satellite visibility conditions described as follows: a)~the percent of time for which at least $1$ Earth-GPS satellite is visible, since this is the minimum number required to estimate the clock bias and drift; b)~the percent time for which at least $4$ Earth-GPS satellites are visible, since this is the minimum number to estimate the full state vector that includes position, clock bias, velocity and clock drift;
    \item Maximum ECOP to identify the region of maximum continuous time when no Earth-GPS satellites are visible; 
    \item Root-Mean-Square~(RMS) errors in clock estimates to analyze the performance of our time-transfer from Earth-GPS for reduced measurement update rate with a sampling period of $\measupdaterate=m\predrate$, where $m=5$; and
    \item Lunar UERE metric that characterize the ranging measurement accuracy of signals transmitted by an LNSS satellite. 
    Recollect from an earlier explanation in Section~\ref{sec:proposed_time_transfer_technique} that the lunar UERE metric depends on the RMS error in clock bias. 
\end{enumerate}


We consider a case study to be desirable if it exhibits either some or all of the following: greatest satellite visibility, least maximum ECOP, and least lunar UERE. 
To perform sensitivity analysis of our case studies, we compute the lunar UERE metric for different reduced measurement update rates with $m=1,5,30,60$. 
Note that, we perform $m=1$ for baseline comparison~(not reduced rate) since it depicts the case where the measurement update step is executed whenever Earth-GPS satellites are visible, and thus depicts the original framework of timing filter proposed in our prior work~\cite{bhamidipati2021design}. 

\subsection{Across Orbit Types: Variation in Satellite Visibility and Maximum ECOP}
\begin{figure}[H]
	\centering
	\begin{subfigure}[b]{0.49\textwidth}
		\includegraphics[width=0.98\textwidth]{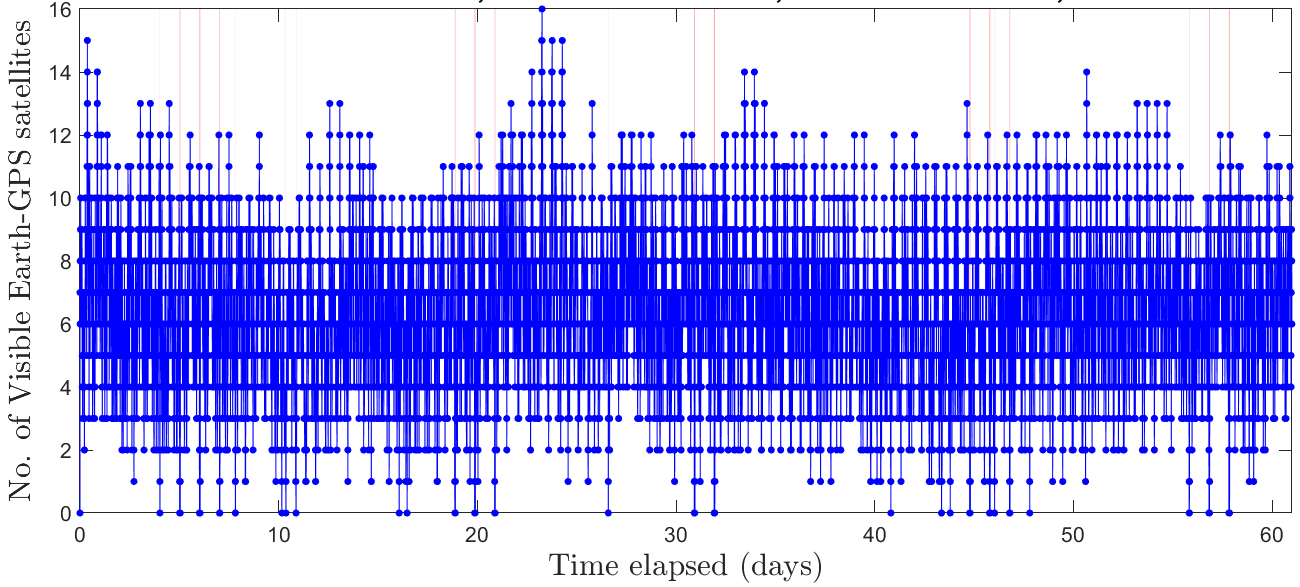}
		\caption{ELFO: Full satellite visibility}
		\label{fig:elfo_fullsatvis}
	\end{subfigure}	
	\hspace{2mm}
	\begin{subfigure}[b]{0.49\textwidth}
		\includegraphics[width=0.98\textwidth]{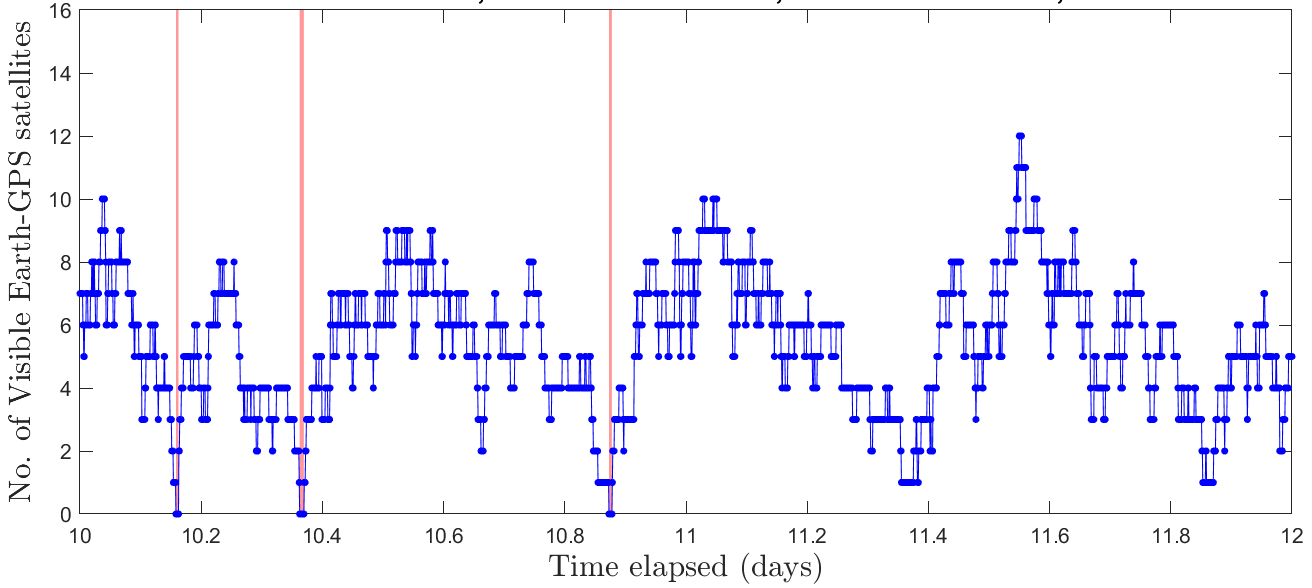}
		\caption{ELFO: Zoomed-in satellite visibility}
		\label{fig:elfo_subsatvis}
	\end{subfigure}
	\vfill
	\vspace{5mm}
	\begin{subfigure}[b]{0.49\textwidth}
		\includegraphics[width=0.98\textwidth]{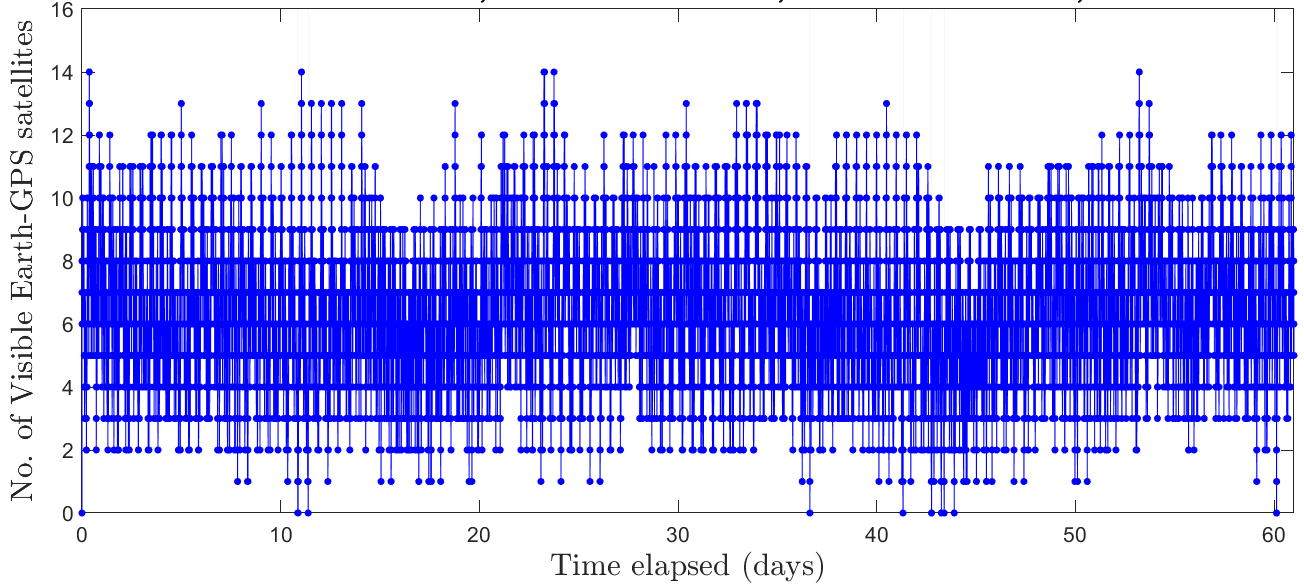}
		\caption{NRHO: Full satellite visibility}
		\label{fig:nrho_fullsatvis}
	\end{subfigure}
	\hspace{1mm}
	\begin{subfigure}[b]{0.49\textwidth}
		\includegraphics[width=0.98\textwidth]{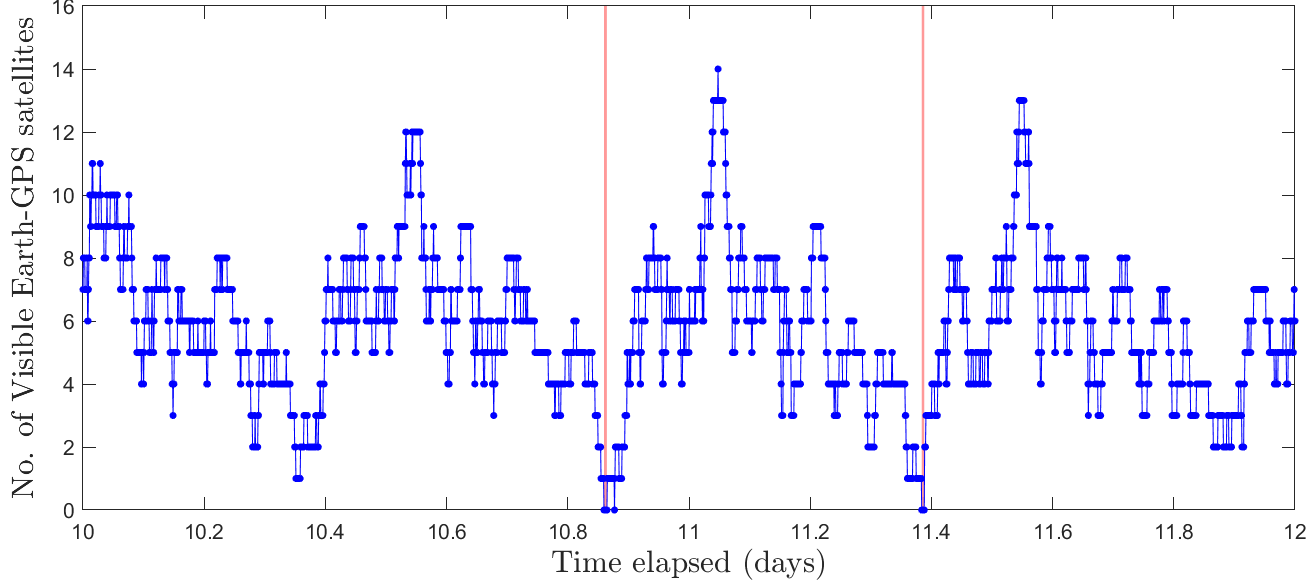}
		\caption{NRHO: Zoomed-in satellite visibility}
		\label{fig:nrho_subsatvis}
	\end{subfigure}	
	\vfill
	\vspace{5mm}
	\begin{subfigure}[b]{0.49\textwidth}
		\includegraphics[width=0.98\textwidth]{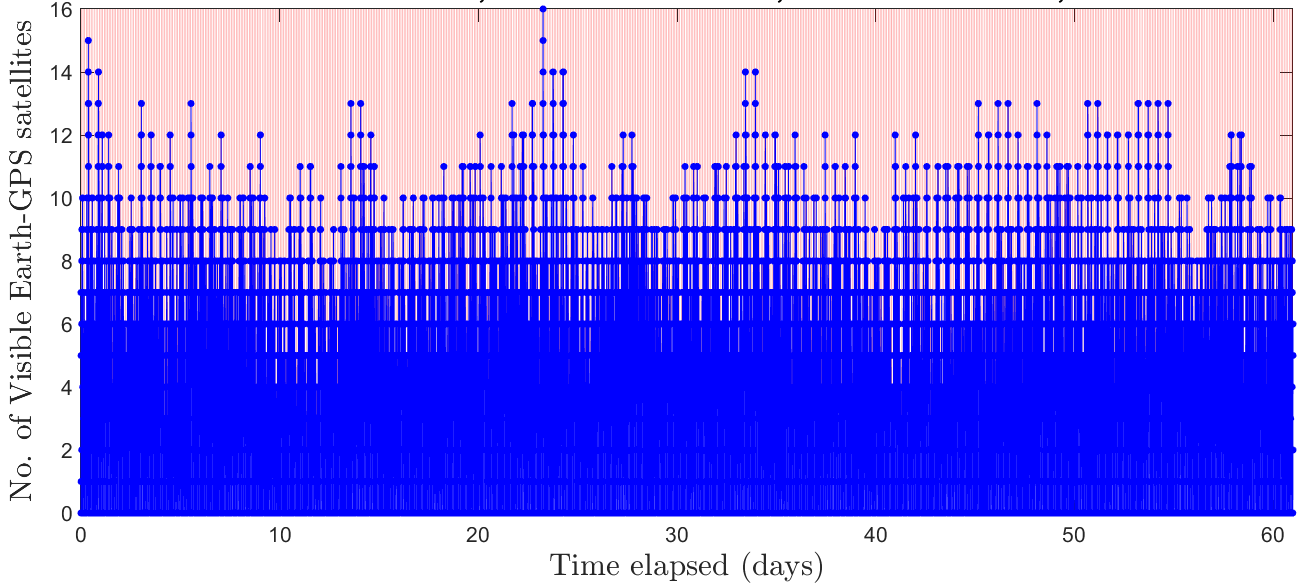}
		\caption{LLO: Full satellite visibility}
		\label{fig:llo_fullsatvis}
	\end{subfigure}
	\hspace{1mm}
	\begin{subfigure}[b]{0.49\textwidth}
		\includegraphics[width=0.98\textwidth]{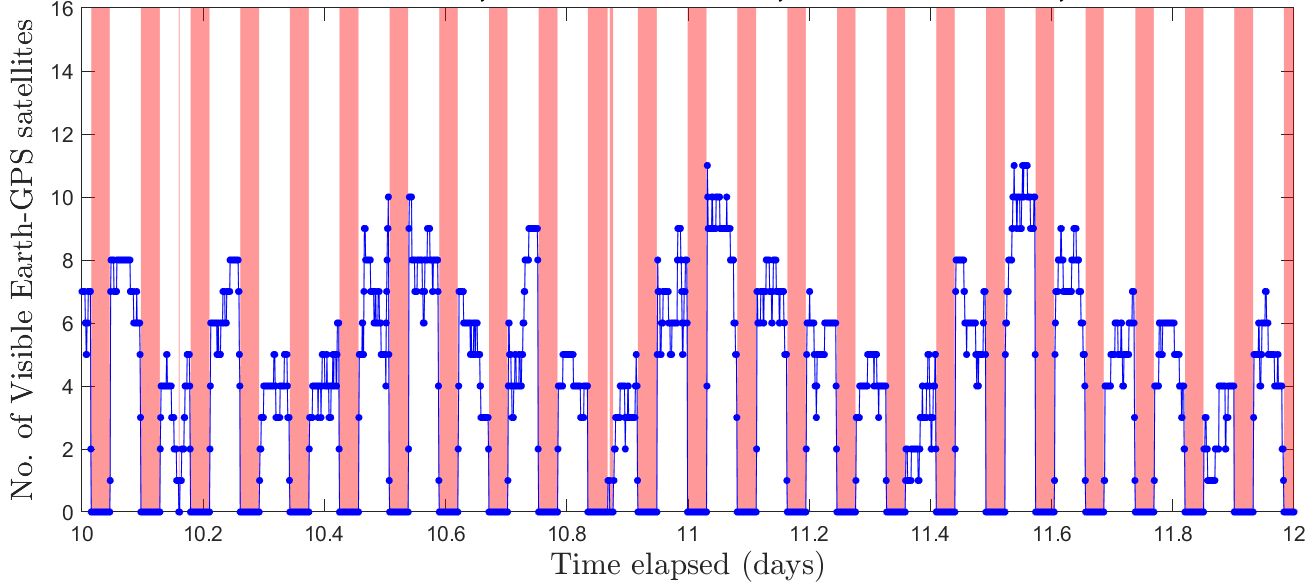}
		\caption{LLO: Zoomed-in satellite visibility}
		\label{fig:llo_subsatvis}
	\end{subfigure}	
	\vfill
	\vspace{5mm}
	\begin{subfigure}[b]{0.49\textwidth}
		\includegraphics[width=0.98\textwidth]{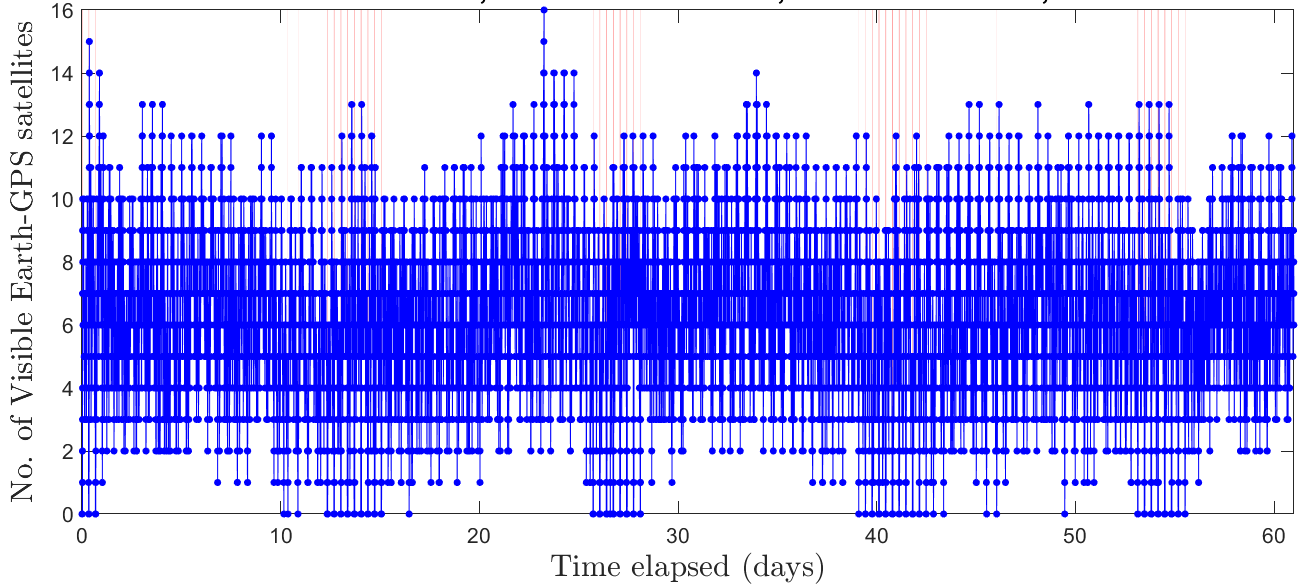}
		\caption{PCO: Full satellite visibility}
		\label{fig:pco_fullsatvis}
	\end{subfigure}
	\hspace{1mm}
	\begin{subfigure}[b]{0.49\textwidth}
		\includegraphics[width=0.98\textwidth]{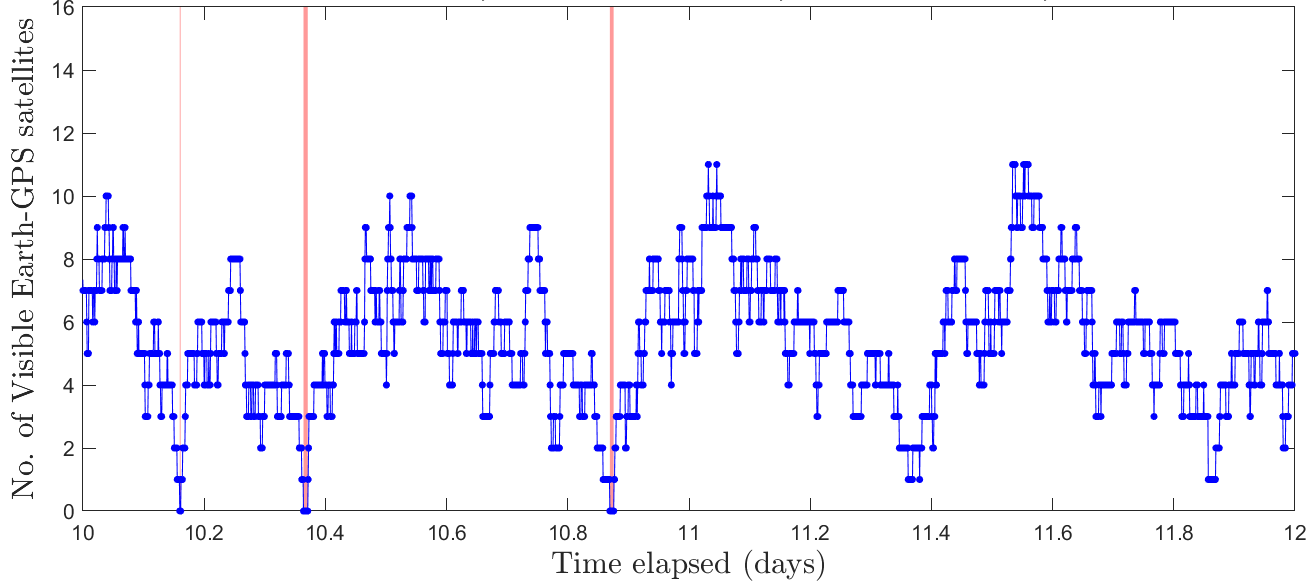}
		\caption{PCO: Zoomed-in satellite visibility}
		\label{fig:pco_subsatvis}
	\end{subfigure}	
	\caption{Earth-GPS satellite visibility and maximum ECOP across different orbit types, where blue dotted lines indicate the number of visible Earth-GPS satellites and the red vertical bars indicate regions of ECOP; 
	a), c), e) and g) show the satellite visibility for the entire time duration in ELFO, NRHO, LLO and PCO, respectively; 
	b), d), f) and h) show the zoomed-in satellite visibility for a shorter time segment of $2$ days. 
	We observe that NRHO not only exhibits a least maximum ECOP of $420$~\si{s} but also the greatest at least one satellite visibility of $99.9\%$.}
	\label{fig:experiment_maximum_timing_errors}
\end{figure}

For the four orbit types considered in our case study, we showcase the number of visible Earth-GPS satellites by blue in Figs.~\ref{fig:elfo_fullsatvis}-\ref{fig:pco_subsatvis} while the highlighted red vertical bars indicate the regions of ECOP.
Based on Table~\ref{table:experiment_ECOP_satvis}, we observe that NRHO achieves the greatest at least one satellite visibility, which is $99.9\%$ of the total time, and also the greatest at least four satellite visibility, which is $92.9\%$ of the total time. 
Furthermore, NRHO also exhibits the least maximum ECOP of only $420$~\si{s}, while the other orbit types experience ECOP of at least $2880$~\si{s}. 
These observations related to NRHO seem reasonable since an LNSS satellite in NRHO operates at high altitudes ranging between $4500$~\si{km}-$700000$~\si{km} above the Moon's surface, and thus experiences fewer occultations from Earth and the Moon. 
Comparing Figs.~\ref{fig:elfo_subsatvis},~\ref{fig:nrho_subsatvis},~\ref{fig:llo_subsatvis} and~\ref{fig:pco_subsatvis}, we observe that LLO, which has a low altitude of $100$~\si{km}, exhibits the smallest time spacing between consecutive regions of ECOP. 
\begin{table}[H]
\centering
\caption{Comparison analysis across different orbit types, where NRHO exhibits the least maximum ECOP as well as the greatest visibility of both at least one and at least four Earth-GPS satellites.}
\begin{tabular}{|c|c|c|c|}
\hline
\multirow{2}{*}{\textbf{\begin{tabular}[c]{@{}c@{}}Orbit Type\\ \end{tabular}}} & \multirow{2}{*}{\textbf{\begin{tabular}[c]{@{}c@{}}Max ECOP (\si{s})\\ \end{tabular}}} & \multicolumn{2}{c|}{\textbf{Satellite Visibility (\%)}} \\ \cline{3-4} 
 &  & \textbf{$\geq 1$} & \textbf{$\geq 4$} \\ \hline
\textbf{ELFO} & 3360 & 99.2 & 92.1 \\ \hline 
\textbf{NRHO} & \textbf{420} & \textbf{99.9} & \textbf{92.9} \\ \hline
\textbf{LLO} & 2880 & 61.5 & 55.6 \\ \hline
\textbf{PCO} & 3900 & 98.1 & 90.8 \\ \hline
\end{tabular}
\label{table:experiment_ECOP_satvis}
\end{table}

\subsection{Across Clock Types: Variation in RMS Timing Errors}
Table~\ref{table:rmse_timing_errors} compares the RMS estimation error in clock bias and drift across different case studies. 
Intuitively, RMS timing error provides insights regarding the component of lunar UERE metric, whose results will be discussed in the next two subsections. 
\begin{table}[H]
\centering
\caption{Comparison analysis across different orbit types, where NRHO exhibits the least maximum ECOP as well as the greatest visibility of both at least one and at least four Earth-GPS satellites.}
\begin{tabular}{|cc|c|c|c|c|}
\hline
\multicolumn{2}{|c|}{\textbf{\begin{tabular}[c]{@{}c@{}}RMS Timing Error \\ in Clock Type\end{tabular}}} & \textbf{ELFO} & \textbf{NRHO} & \textbf{LLO} & \textbf{PCO} \\ \hline
\multicolumn{1}{|c|}{\multirow{2}{*}{\textbf{\begin{tabular}[c]{@{}c@{}}Microchip\\ CSAC\end{tabular}}}} & \textbf{Bias (\si{\mu s})} & 0.0405 & 0.0383 & 0.0679 & 0.0408 \\ \cline{2-6} 
\multicolumn{1}{|c|}{} & \textbf{Drift (\si{ns/s})} & 0.0407 & 0.0401 & 0.0419 & 0.0404 \\ \hline

\multicolumn{1}{|c|}{\multirow{2}{*}{\textbf{\begin{tabular}[c]{@{}c@{}}Microchip \\ MAC\end{tabular}}}} & \textbf{Bias (\si{\mu s})} & 0.0213 & 0.0215 & 0.0278 & 0.0217 \\ \cline{2-6} 
\multicolumn{1}{|c|}{} & \textbf{Drift (\si{ns/s})} & 0.0081 & 0.0081 & 0.0081 & 0.0083 \\ \hline

\multicolumn{1}{|c|}{\multirow{2}{*}{\textbf{\begin{tabular}[c]{@{}c@{}}SRS\\ PRS-10\end{tabular}}}} & \textbf{Bias (\si{\mu s})} & 0.0176 & 0.0177 & 0.0213 & 0.0178 \\ \cline{2-6} 
\multicolumn{1}{|c|}{} & \textbf{Drift (\si{ns/s})} & 0.0045 & 0.0046 & 0.0045 & 0.0046 \\ \hline

\multicolumn{1}{|c|}{\multirow{2}{*}{\textbf{\begin{tabular}[c]{@{}c@{}}Excelitas\\ RAFS\end{tabular}}}} & \textbf{Bias (\si{\mu s})} & 0.0123 & 0.0121 & 0.0143 & 0.0121 \\ \cline{2-6} 
\multicolumn{1}{|c|}{} & \textbf{Drift (\si{ns/s})} & 0.0015 & 0.0015 & 0.0015 & 0.0015 \\ \hline

\multicolumn{1}{|c|}{\multirow{2}{*}{\textbf{\begin{tabular}[c]{@{}c@{}}NASA's\\ DSAC\end{tabular}}}} & \textbf{Bias (\si{\mu s})} & \textbf{0.0044} & \textbf{0.0044} & \textbf{0.0047} & \textbf{0.0037} \\ \cline{2-6} 
\multicolumn{1}{|c|}{} & \textbf{Drift (\si{ns/s})} & \textbf{6.4347$\times 10^{-5}$} & \textbf{6.5224$\times 10^{-5}$} & \textbf{6.2655$\times 10^{-5}$} & \textbf{5.6112$\times 10^{-5}$} \\ \hline
\end{tabular}
\label{table:rmse_timing_errors}
\end{table}

\begin{figure}[H]
	\centering
	\begin{subfigure}[b]{0.49\textwidth}
		\includegraphics[width=0.98\textwidth]{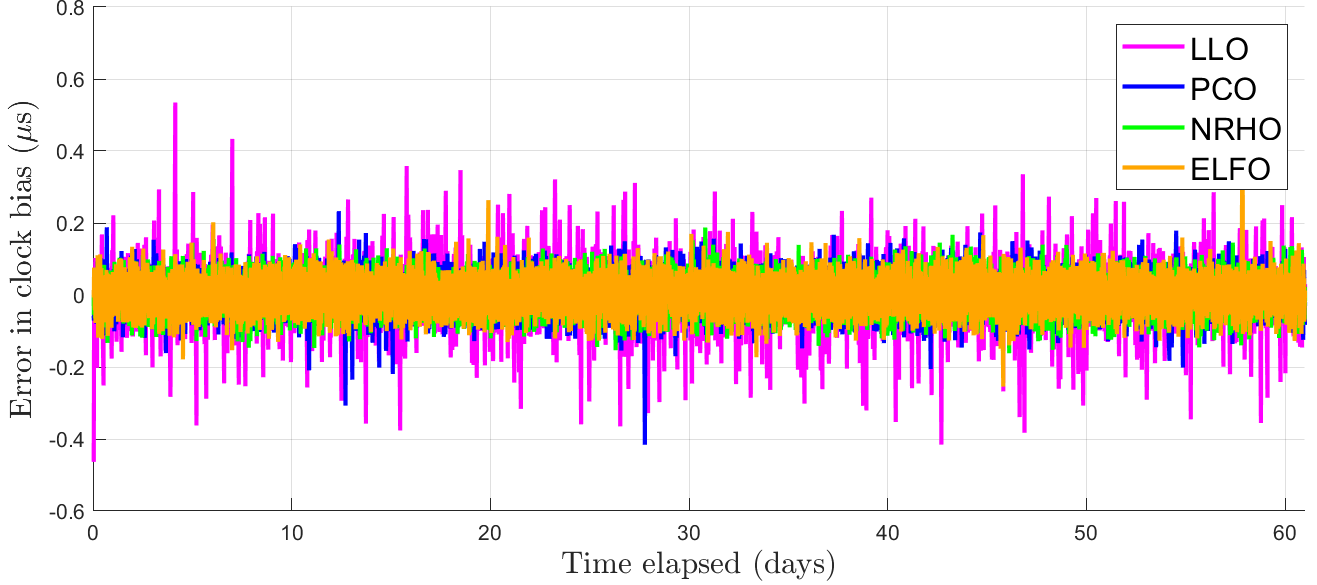}
		\caption{Clock bias error for entire experiment duration}
	\end{subfigure}	
	\hspace{2mm}
	\begin{subfigure}[b]{0.49\textwidth}
		\includegraphics[width=0.98\textwidth]{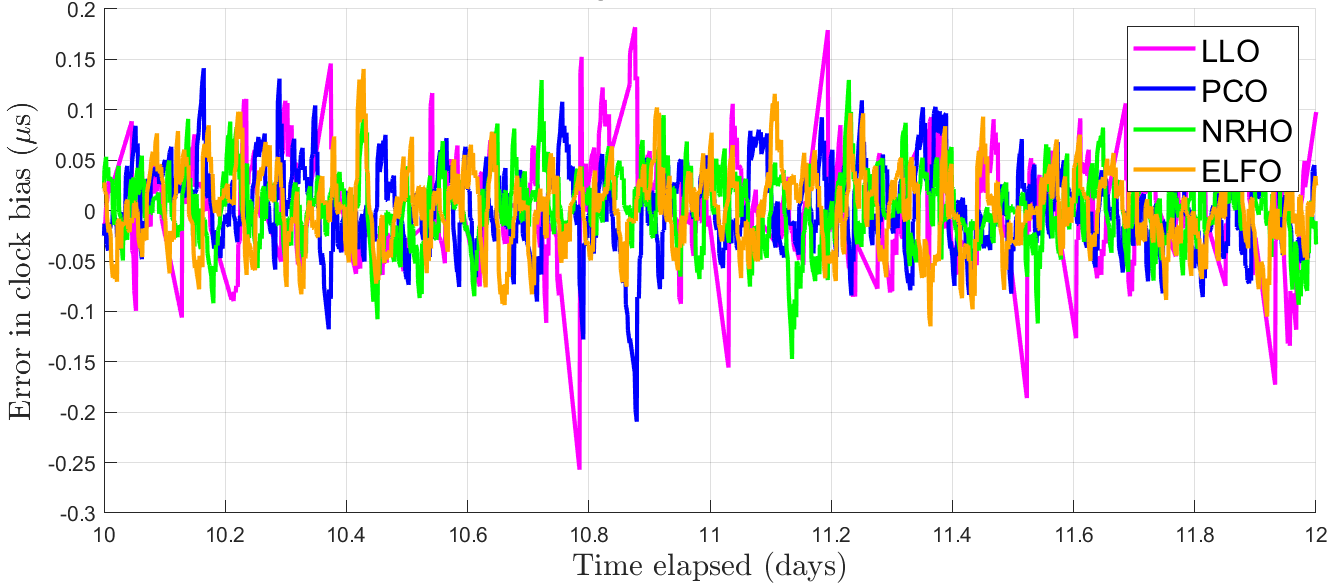}
		\caption{Clock bias error for a smaller time segment}
	\end{subfigure}
	\vfill
	\vspace{5mm}
	\begin{subfigure}[b]{0.49\textwidth}
		\includegraphics[width=0.98\textwidth]{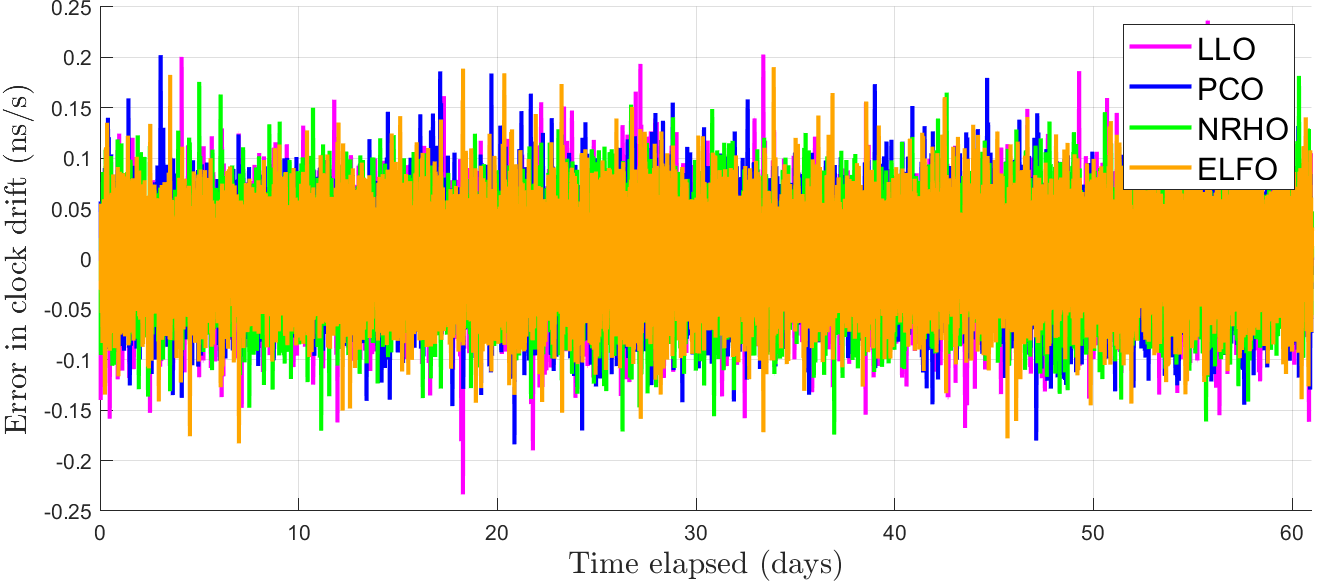}
		\caption{Clock drift error for entire duration}
	\end{subfigure}
	\hspace{1mm}
	\begin{subfigure}[b]{0.49\textwidth}
		\includegraphics[width=0.98\textwidth]{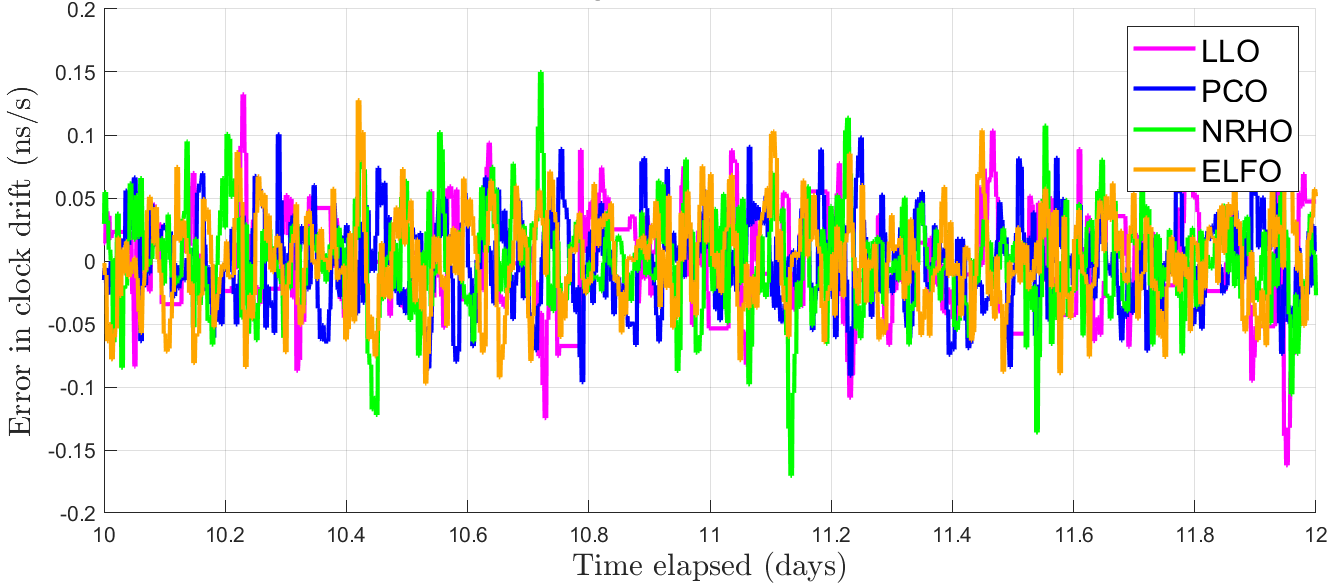}
		\caption{Clock drift error for a smaller time segment}
	\end{subfigure}	
	\caption{Comparison of estimation errors in clock bias and drift across different orbit types with an onboard Microchip CSAC, where ELFO is indicated in orange, NRHO in green, LLO in magenta, and PCO in blue; a) and c) demonstrate the errors in clock bias and clock drift for the entire experiment duration, respectively, while b) and d) demonstrate the zoomed-in errors in clock bias and clock drift for a smaller time segment of $2$~days. We observe that the three orbit types, namely ELFO, NRHO and PCO, demonstrate a comparable RMS timing error of $<0.0408$~\si{\mu s} in clock bias and $<0.0407$~\si{ns/s} in clock drift, while the LLO exhibits slightly higher RMS error in clock bias of $0.0679$~\si{\mu s}.}
	\label{fig:experiment_rmse_timing_errors}
\end{figure}

For a given Microchip CSAC~$(17$~\si{cm^3}$\cdot 0.0035$~\si{kg}$\cdot 0.12$~\si{W}$)$, whose low SWaP characteristics were reported earlier in Table~\ref{table:LNSSsatclocks}, Fig.~\ref{fig:experiment_rmse_timing_errors} provides an illustration of the variation in timing errors across the four orbit types, namely ELFO, NRHO, LLO and PCO. 
In Fig.~\ref{fig:experiment_rmse_timing_errors}, we follow the same color coding as that in Figs.~\ref{fig:LNSSsatelliteThreeOrbitTypes} and~\ref{fig:LNSSsatelliteNRHO} to denote different orbit types, with ELFO indicated in orange, NRHO in green, LLO in magenta and PCO in blue. 
As described previously in the description of validation metrics, we reduce the measurement update rate by setting the sampling period to $\measupdaterate=m\predrate$, where $m=5$.
We observe that the three orbit types, namely ELFO, NRHO and PCO, demonstrate a comparable RMS timing error of $<0.0408$~\si{\mu s} in clock bias and $<0.0407$~\si{ns/s} in clock drift, while the case study based on LLO shows a higher RMS error in clock bias of $0.0679$~\si{\mu s}.
This observation implies that the RMS timing error not only depends on the least maximum ECOP and the greatest satellite visibility but also on orbital parameters, namely eccentricity, inclination and altitude, that govern the geometric configuration between the Earth-GPS and LNSS.
From Table~\ref{table:rmse_timing_errors}, we also observe that, as the SWaP and the timing stability of the onboard clock increases, the variation in RMS error across orbit types becomes less significant, wherein the RMS error in clock bias and drift for DSAC are $<0.0044$~\si{\mu s} and $<6.5224\times 10^{-5}$~\si{ns/s}, respectively.

\subsection{Across Case Studies and Earth-GPS Measurement Update Rates: Sensitivity Analysis of Lunar UERE}
\begin{figure}[H]
	\centering
	\begin{subfigure}[b]{0.43\textwidth}
		\includegraphics[width=0.98\textwidth]{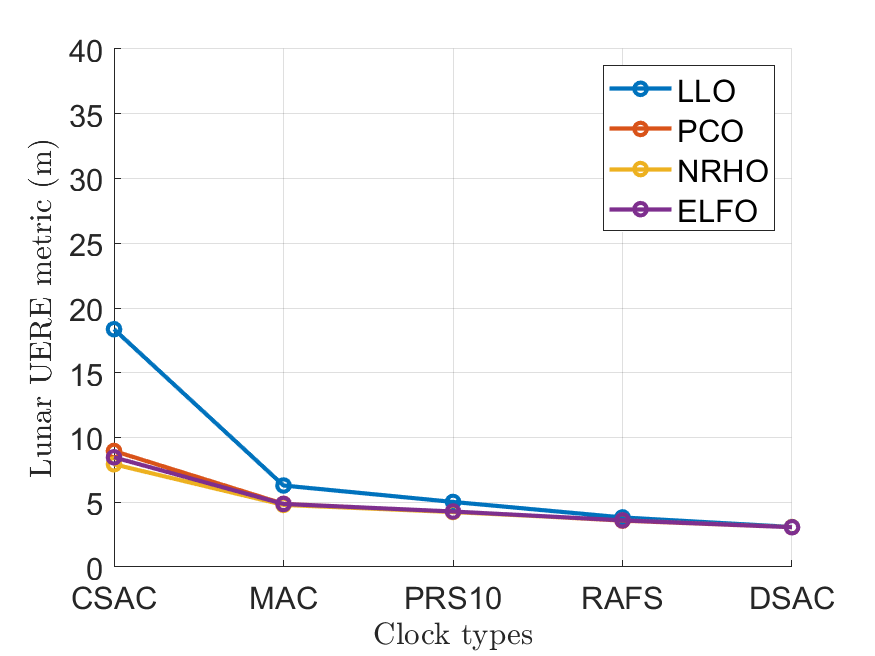}
		\caption{$m=1$ or $\measupdaterate=1$~\si{min}}
		\label{fig:lunarUEREm1}
	\end{subfigure}
	\hspace{2mm}
	\begin{subfigure}[b]{0.43\textwidth}
		\includegraphics[width=0.98\textwidth]{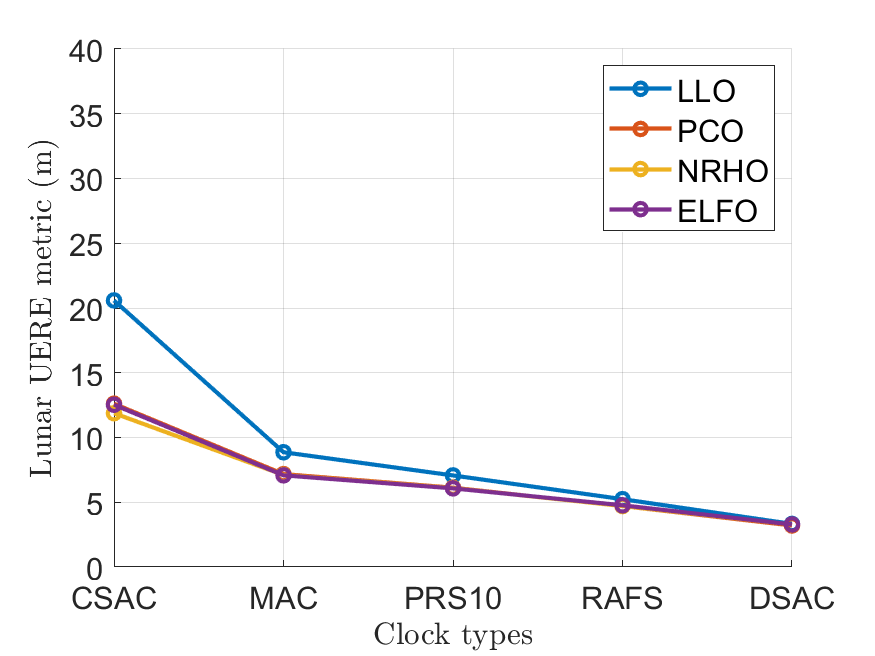}
		\caption{$m=5$ or $\measupdaterate=5$~\si{min}}
		\label{fig:lunarUEREm5}
	\end{subfigure}
	\begin{subfigure}[b]{0.43\textwidth}
		\includegraphics[width=0.98\textwidth]{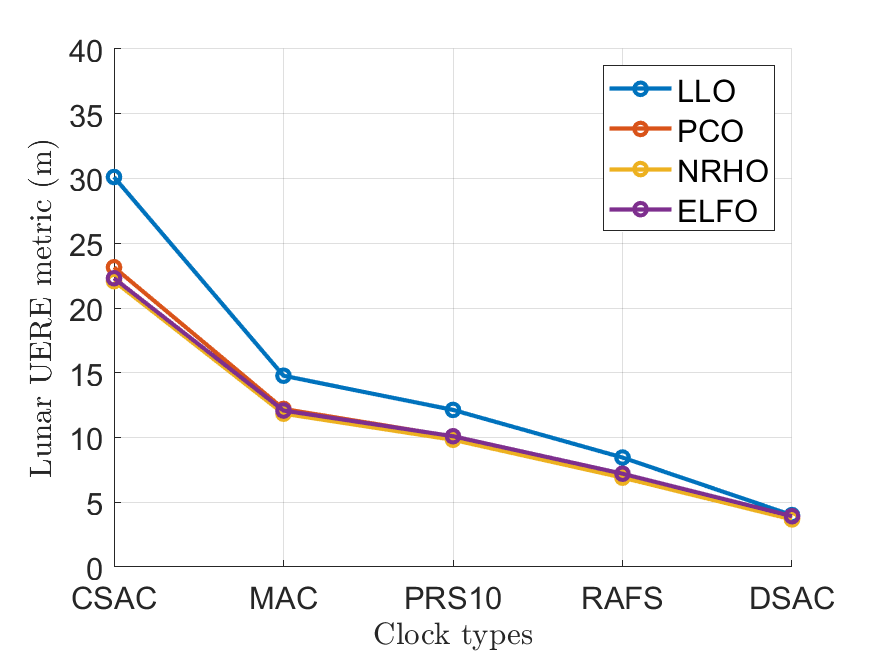}
		\caption{$m=30$ or $\measupdaterate=30$~\si{min}}
		\label{fig:lunarUEREm30}
	\end{subfigure}
	\hspace{2mm}
	\begin{subfigure}[b]{0.43\textwidth}
		\includegraphics[width=0.98\textwidth]{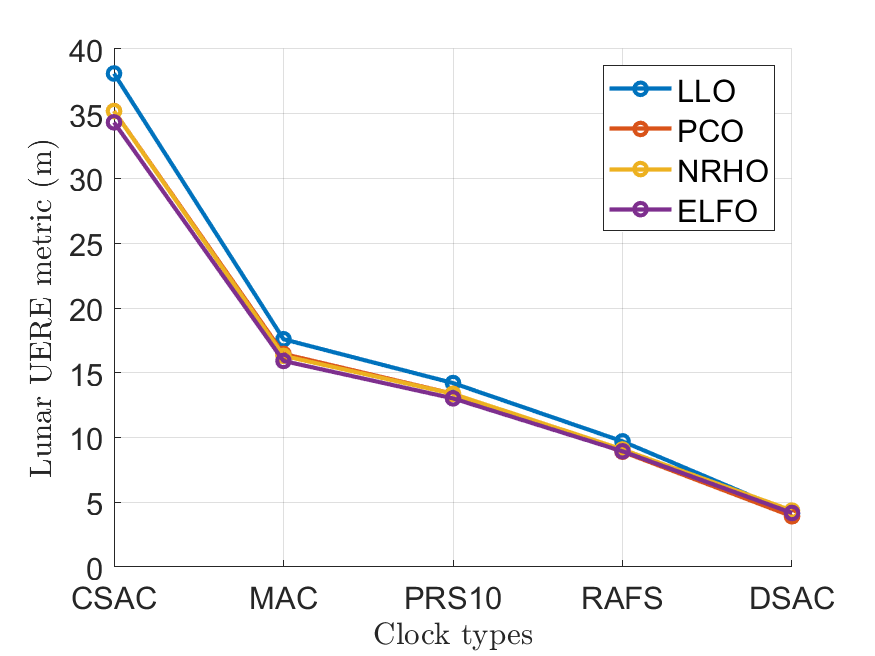}
		\caption{$m=60$ or $\measupdaterate=60$~\si{min}}
		\label{fig:lunarUEREm60}
	\end{subfigure}	
	\caption{Sensitivity analysis of lunar UERE metric across different case studies for reduced measurement update rates where (a)~$m=1$ is the baseline case with no reduction of the measurement update rate, i.e., $\measupdaterate=\predrate$; (b)~$m=5$; (c)~$m=30$; and (d)~$m=60$.} 
\end{figure}

Fig.~\ref{fig:lunarUEREm1} demonstrates the variation in lunar UERE metric across case studies conducted while considering no reduction in the measurement update rate, i.e., we set $\measupdaterate=\predrate=60$~\si{s} or $m=1$. 
We validate that our proposed time-transfer achieves a low lunar UERE of $<10$~\si{m} for most case studies except the one involving the LLO and the Microchip CSAC, which exhibits a value of $18.4$~\si{m}.
These observations imply that, if the desired lunar UERE to be maintained by an LNSS satellite is, for instance, $<10$~\si{m} at all times, we can easily opt for an onboard clock that falls in the lower end of the SWaP spectrum, such as Microchip CSAC~$(17$~\si{cm^3}$\cdot 0.0035$~\si{kg}$\cdot 0.12$~\si{W}$)$ or Microchip MAC~$(50$~\si{cm^3}$\cdot 0.0084$~\si{kg}$\cdot 0.5$~\si{W}$)$, instead of a high-SWaP clock, such as Excelitas RAFS or NASA's DSAC.
Given a desired lunar UERE to maintain, we can also wisely choose the orbit type that is easy to maintain and has a longer lifespan, such as LLO, PCO or ELFO, over the more complex NRHO that requires constant station-keeping maneuvers to maintain stability.   
Furthermore, for future investigations, we observe the potential for a heterogenous, SmallSat-based lunar PNT constellation, wherein we can choose the grade of the onboard clock based on the orbit of each LNSS satellite to satisfy a desired lunar UERE.
For instance, to maintain a desired lunar UERE of $<5$~\si{m}, we observe from Fig.~\ref{fig:lunarUEREm1} that we can design a heterogenous LNSS constellation based in ELFO and LLO, wherein the satellites in LLO can be equipped with a PRS-10 clock while the ones in ELFO can be equipped with a lower-SWaP Microchip MAC.  

Figures~\ref{fig:lunarUEREm5}-\ref{fig:lunarUEREm60} show the variation in lunar UERE metric for three cases of reduced update rates, where $\measupdaterate=m\predrate$ with $m=5$, $m=30$ and $m=60$ respectively.
As the Earth-GPS measurement update rate increases for low-SWaP clocks with sampling periods from $\measupdaterate=60$~\si{min}~($m=60$) to $\measupdaterate=1$~\si{min}~($m=1$), we observe an increased sensitivity of the lunar UERE metric across lunar orbit types, i.e., difference in value between LLO and others increases. 
We also demonstrate that the estimated lunar UERE is $<30$~\si{m} for a reduced Earth-GPS measurement update rate with a sampling period of up to $\measupdaterate=30$~\si{min}, which is comparable in order of magnitude to that of the baseline case with $m=1$~(in Fig.~\ref{fig:lunarUEREm1}) as well as the legacy Earth-GPS. 
Thus, we validate that our LNSS design that utilizes time-transfer from Earth-GPS at a reduced measurement update rate to lower the SWaP requirements of the onboard clock. 
Furthermore, we can re-observe the potential for heterogeneous, SmallSat-based PNT constellation explained above, wherein we can intelligently choose not only the grade of the onboard clock but also the reduced update rate of the timing filter based on the LNSS satellite orbit so as to satisfy a desired lunar UERE. 
\section{Conclusion}\label{sec:conclusion}
We performed an exhaustive case study analysis for designing a SmallSat-based Lunar Navigation Satellite System~(LNSS) with time-transfer from Earth-GPS, wherein we investigate the trade-off between different design considerations related to the onboard clock and the lunar orbit type.
In our time-transfer, we alleviated the Size, Weight and Power~(SWaP) requirements of the onboard clocks by leveraging the intermittently available Earth-GPS signals to provide timing corrections, and also designed a lunar User Equivalent Range Error~(UERE) metric to characterize the ranging accuracy of the LNSS satellite. 

Using high-fidelity simulations of an LNSS satellite in the System Tool Kit~(STK) software by Analytical Graphics, Inc.~(AGI), we designed multiple case studies comprising five onboard clocks and four lunar orbit types. 
We validated that the least maximum Earth-GPS Continual Outage Period~(ECOP) of only $420$~\si{s} is observed for a Near-Rectilinear Halo Orbit~(NRHO) since it experiences fewer occultations from Earth and the Moon given its high altitude.
We demonstrated that a low lunar UERE of $<30$~\si{m} can be achieved using low-SWaP onboard clocks (e.g., Microchip chip scale atomic clock, Stanford Research Systems PRS-10) even for reduced Earth-GPS measurement update rates with sampling period of up to $30$~\si{min}. 
Through a case study analysis of our time-transfer from Earth-GPS, we demonstrated that lower-SWaP onboard clocks and easier-to-maintain lunar orbit types can be chosen over the others to still achieve a desired lunar UERE across the entire LNSS constellation. 

\section*{ACKNOWLEDGEMENTS}
We would like to thank the Analytical Graphics, Inc. (AGI) Educational Alliance Program (EAP) for providing us with the System Tool Kit (STK) software license to perform this research work. 
We would also like to thank Keidai Ilyama for the insightful discussions related to this work and Daniel Neamati for reviewing this paper. 

\bibliographystyle{ION_bibstyle}
\bibliography{references}

\end{document}